\documentclass[runningheads]{llncs}
\usepackage[T1]{fontenc}
\usepackage{graphicx,verbatim}

\usepackage{amssymb}
\usepackage{latexsym}
\usepackage{framed,multirow}
\usepackage{amsmath}

\usepackage{pifont}
\usepackage{tabularx} 
\usepackage{array}

\usepackage{hyperref}

\newcommand{\cmark}{\ding{51}} 
\newcommand{\xmark}{\ding{55}} 
\usepackage{color}

\begin{document}
\title{HypOProto: Hyperbolic Ordinal Prototypes for Left Ventricular Filling Pressure Classification}
%

\author{Victoria Wu\inst{1} \and Nima Hashemi\inst{1} \and Hooman Vaseli\inst{1} \and Christina Luong\inst{2} \and Purang Abolmaesumi\inst{1} \and Teresa S. M. Tsang\inst{2}}  
\authorrunning{V. Wu et al.}


\institute{Department of Electrical and Computer Engineering, The University of British Columbia, Vancouver, BC, Canada \\ 
\and Vancouver General Hospital, Vancouver, BC, Canada \\
\email{purang@ece.ubc.ca}
\footnote{T. S.M. Tsang and P. Abolmaesumi are joint senior authors.}
}
  
\maketitle              
\begin{abstract}
Echocardiography (echo) is a widely used imaging modality for assessing cardiac function, with Left Ventricular Filling Pressure (LVFP) serving as a critical physiological marker for conditions such as heart failure. Standard LVFP classification into normal \emph{vs} elevated categories relies on the Doppler-derived $E/e'$ ratio, which is operator-dependent and often unavailable in resource-limited settings, motivating methods that infer LVFP directly from B-mode echo. Existing deep learning approaches achieve high performance but remain largely black-box, limiting clinical interpretability. We propose HypOProto, a hyperbolic, ordinal prototype-based framework for interpretable LVFP classification using a frozen, explainable foundation model backbone. HypOProto arranges prototypes along the physiological $E/e'$ scale, placing borderline cases near the hyperboloid root where small angular differences separate similar cases, while normal and elevated cases occupy outward positions reflecting increasing diagnostic certainty. This hyperbolic geometry encodes clinically meaningful ordinal relationships and improves interpretability. We also introduce a novel Hyperbolic Prototype Angular Separation (HyperPAS) loss, enforcing inter-class prototype separation in hyperbolic space. HypOProto achieves SOTA performance while maintaining transparency, and highlights clinically relevant regions in visualizations. This work represents the first prototype-based framework for LVFP classification in echo. Our code can be found at https://github.com/DeepRCL/HypOProto.


\end{abstract}
\section{Introduction}

Echocardiography (echo) is the most widely used imaging modality for evaluating cardiac function due to its non-invasive nature, portability, and ability to visualize the heart in real time. Beyond anatomical assessment, echo provides functional information that is critical for diagnosing diastolic dysfunction. Among these functional markers, Left Ventricular Filling Pressure (LVFP) is a primary indicator across various cardiac pathologies including heart failure~\cite{ase2025diastolic}. Accurate estimation of LVFP is therefore essential in both clinical and research settings, as it directly informs diagnosis, prognosis, and treatment planning.
In clinical practice, LVFP is typically estimated using Doppler echocardiography, particularly through the ratio of mitral inflow velocity (E) to the early diastolic mitral annular velocity (e'). This $E/e'$ ratio is then typically used to classify patients as having normal or elevated LVFP~\cite{ase2025diastolic}. While effective, the measurement is highly operator-dependent, sensitive to image quality and acquisition angle, and often unavailable in resource-limited or point-of-care settings~\cite{minners2010inconsistent}. These limitations motivate the exploration of inferring LVFP directly from B-mode echo images, which are more consistent and widely accessible.

Despite a growing number of works on using deep learning (DL) for B-mode echo analysis, there are currently no methods explicitly targeting LVFP estimation. The feasibility of this task has been indirectly demonstrated by Akerman et al.~\cite{akerman2023automated}, who successfully diagnosed heart failure with preserved ejection fraction (HFpEF) from apical four-chamber (A4C) videos alone. Given that LVFP is the primary physiological signal for HFpEF, this result implies that B-mode echo encodes sufficient information for learning LVFP. 

Most prior DL approaches in echocardiography are black-box models that achieve high performance but limited transparency, which hinders clinical adoption. Prototype-based networks offer an explainable AI (XAI) alternative by providing case-based reasoning in cardiac imaging~\cite{ghamary2025protoefnet,vaseli2023protoasnet}. However, these interpretable methods have not been applied to LVFP estimation.


A further challenge in scaling DL to large echo datasets is the reliance on fully supervised, end-to-end training, which proves impractical given limited labeled data. EchoPrime~\cite{vukadinovic2024echoprime} addressed this by providing robust foundation model representations through self-supervised pretraining on large-scale echo data. However, its pooling architecture remains black-box and limits explainability. 


Recent works have demonstrated that hyperbolic geometry improves performance in medical image classification by embedding naturally hierarchical relationships~\cite{gonzalez2025hyperbolic,huang2025hyperpath,vaseli2025hiprotonet}. In hyperbolic space, there is a characteristic training tendency in which more ambiguous samples remain closer to the manifold origin (root), while more distinctive and class-specific patterns move outward along curved branches toward the boundary, preserving semantic hierarchical structure without Euclidean distortions~\cite{hu2024hyperbolic,khrulkov2020hyperbolic}. This behavior is well suited to LVFP classification. Although typically framed as a binary task (normal \emph{vs.}\ elevated), LVFP is inherently ordinal, governed by the continuous E/e$^\prime$ ratio with a clinical threshold at $14$. Cases near this boundary are often more ambiguous, and many elevated studies cluster just above the cutoff, making them particularly difficult to distinguish. Placing these cases closer to the hyperbolic root allows the model to allocate more angular capacity for separating subtle directional differences.

We propose \textbf{HypOProto}, a hyperbolic, ordinal prototype-based framework for interpretable LVFP classification from B-mode echo using a frozen, explainable patch-based foundation model backbone. HypOProto arranges prototypes ordinally along the physiological E/e$^\prime$ scale, placing borderline and difficult cases where $E/e' \approx 14$ near the hyperboloid root, where small angular differences capture subtle distinctions, while clearly normal cases and cases with very high $E/e'$ move outward with increasing diagnostic certainty (Figure~\ref{fig:network}b). This geometry encodes clinically meaningful ordinal structure and improves interpretability. Our key contributions are: (1) the first prototype-based echocardiography framework with a frozen foundation-model backbone and the first interpretable model for LVFP classification, (2) a hyperbolic ordinality mechanism that organizes prototypes according to the $E/e'$ scale to better handle ambiguous cases, and (3) a novel Hyperbolic Prototype Angular Separation (HyperPAS) loss that promotes discriminative prototype directions in hyperbolic space.

\section{Method}
\begin{figure}[t]
\centering
\includegraphics[width=0.99\textwidth]{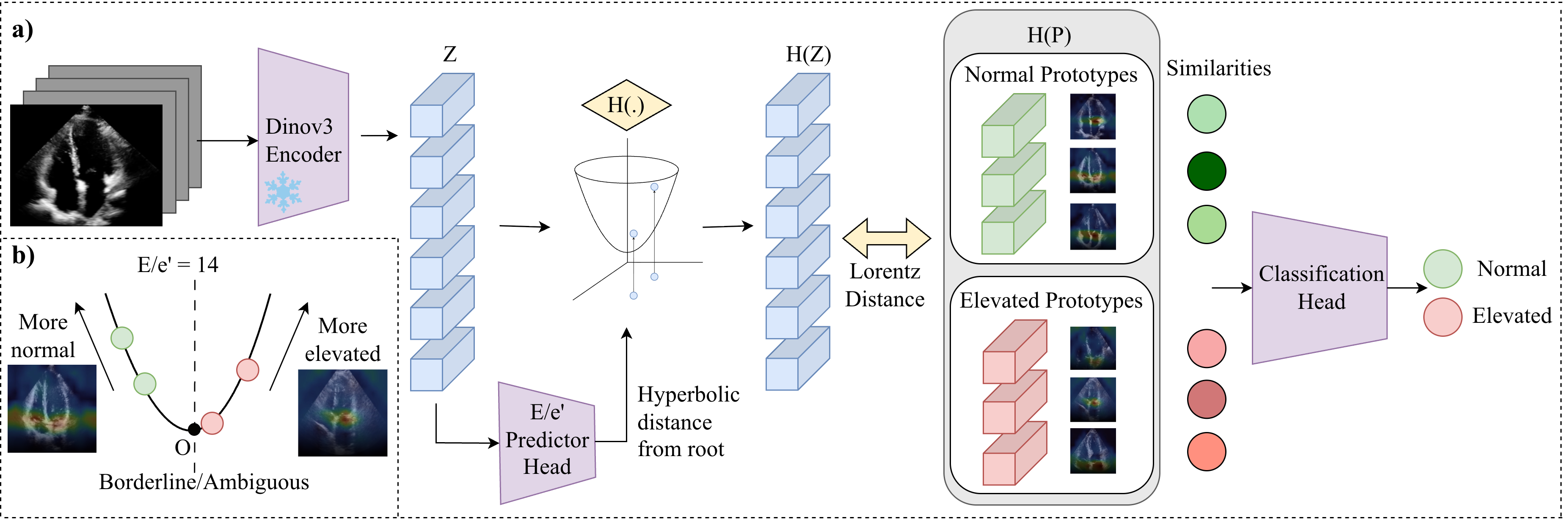}
\caption{a) Method overview: Videos are encoded with DINOv3 to obtain patch features, then passed to an $E/e'$ prediction head, which determines the hyperbolic radius for exponential mapping. Resulting embeddings are compared with hyperbolic prototypes for classification. b) Ordinal structure of the hyperbolic space: cases near the diagnostic threshold lie close to the root, while more confidently normal and elevated cases are positioned farther away.}
\label{fig:network}
\end{figure}

Our goal is to classify LVFP by comparing visual features with a fixed set of learnable prototypes (Figure~\ref{fig:network}a). 
This is an inherently ordinal task and to capture this structure, both features and prototypes are projected into hyperbolic space, where radial distance from the origin reflects proximity to the clinical cutoff.
Each prototype has a class label and a learnable $E/e'$ value for hyperbolic projection, and all prototypes are jointly optimized with the feature embeddings.

\subsubsection{Feature Extraction.}
Unlike prior prototypical approaches that train the backbone end-to-end~\cite{ghamary2025protoefnet,vaseli2023protoasnet,vaseli2025hiprotonet}, we use a frozen pretrained DINOv3 encoder~\cite{simeoni2025dinov3} trained on ~10,000 multi-view echo studies to generate clip-level embeddings. For a video $x \in \mathbb{R}^{T_0 \times H_0 \times W_0 \times 3}$, the extractor outputs $F \in \mathbb{R}^{T \times H \times W \times D}$, computed once and fixed during training. This design eliminates the computational cost of backbone optimization.
We then apply spatiotemporal attention over the embeddings to identify $k$ salient regions of interest (ROIs), $A$, that guide prototype assignment. The attended representations are obtained via the element-wise product $Z = A \odot F$.

\subsubsection{Hyperbolic Ordinality Projection.}
To capture the inherent ordinality of LVFP, we project features into a Lorentzian hyperbolic space where the radial distance from the origin reflects proximity to the clinical $E/e'$ cutoff~\cite{law2019lorentzian}. Borderline cases lie near the root of the hyperboloid, while clearly normal or elevated cases are mapped farther away, enabling the model to encode severity information directly in the feature geometry.

We use a lightweight prediction head to estimate a scalar $\hat{E}$ corresponding to the $E/e'$ value. This estimate is converted to a hyperbolic radius $r$ using a smooth distance function centered at the clinical cutoff:
\begin{equation}
r = r_{\min} + (r_{\max} - r_{\min}) \tanh\!\left(\frac{\alpha \sqrt{(\hat{E} - 14)^2 + \epsilon^2}}{r_{\max}}\right),
 \label{eq:r}
\end{equation}
where $r_{\min} > 0$ ensures numerical stability, $r_{\max}$ bounds the radius, $\alpha$ controls growth away from the decision boundary, and $\epsilon$ is a smoothing factor that preserves gradients near the cutoff.

Following~\cite{desai2023hyperbolic}, we represent each Euclidean representation ${Z}$ in the Lorentz hyperboloid as: 
\begin{equation}
H(Z) = [Z_{\text{space}}, Z_{\text{time}}].
\end{equation}
Features are lifted to the hyperboloid using an exponential map at the origin with the learned radius $r$ that encodes ordinal information based on the predicted $E/e'$ value (Eq. ~\ref{eq:r}). Let $\mathbf{u} = \frac{{Z}}{\|{Z}\|_2}$ denote the unit vector in the direction of $Z$. The hyperbolic space components are then
\begin{equation}
{Z}_{\text{space}} = \frac{\sinh(\sqrt{c}\,r)}{\sqrt{c}}\,{u}, \quad
Z_{\text{time}} = \sqrt{1/c + \|{Z}_{\text{space}}\|_2^2},
\end{equation}
where $c$ is the hyperboloid curvature. This preserves angular information from the Euclidean features while encoding ordinality through the radial coordinate.

\subsubsection{Prototype Classification.}
Along with the feature representations $Z$, each prototype $P$ is projected to the Lorentz hyperboloid using the same procedure, with the radius determined from the prototype's $E/e'$ value. For a given sample, we compute pairwise geodesic distances between its hyperbolic embeddings ${H}({Z_{k}})$ and all prototype embeddings $\{{H}({P}_k)\}_{k=1}^K$ based on the Lorentzian inner product:
\begin{equation}
d({H}({Z}_k), {H}({P}_k)) = \frac{1}{\sqrt{c}} \operatorname{arccosh}\Big(-c \,\langle {H}({Z}_k), {H}({P}_k) \rangle_\mathcal{L}\Big).
\end{equation}
These distances are then converted to similarities, which allows the model to perform classification by directly comparing sample embeddings to prototypes while respecting the ordinal structure encoded in hyperbolic space.

Prototypes are periodically projected onto the nearest training features in hyperbolic space. Specifically, each prototype embedding is replaced by the hyperbolic representation of the closest sample belonging to the same class and with an $E/e'$ value within a tolerance $\Delta_l$ of the prototype’s current value. This enables direct visualization while preserving the hyperbolic structure.

\subsubsection{Hyperbolic Optimization Objectives.}

Our training objective combines complementary losses to promote accurate classification, structured prototypes, and coherent attention. We use cross-entropy for classification, hyperbolic prototype cluster and separation losses, and standard $\ell_1$ and transformation-consistency regularization (see~\cite{vaseli2025happi} for details). We also apply a mean absolute error (MAE) loss to the predicted $E/e'$ values.

In addition, we introduce a novel {Hyperbolic Prototype Angular Separation (HyperPAS) loss}, adapted from~\cite{ghamary2025protoefnet}. This loss encourages prototypes of the same class to align in similar directions while pushing prototypes of different classes apart. Specifically, prototypes $P_i$ are log-mapped to the tangent space at the origin and normalized. For each prototype, an InfoNCE-style angular objective separates same-class from different-class prototypes:

\begin{equation}
\mathcal{L}_i = - \log \frac{\sum_{j \in \mathcal{P}_i^+} \exp\Big(\frac{\mathrm{sim}(\hat{p}_i, \hat{p}_j)}{\tau}\Big)}
{\sum_{j \in \mathcal{P}_i^+} \exp\Big(\frac{\mathrm{sim}(\hat{p}_i, \hat{p}_j)}{\tau}\Big) + \sum_{k \in \mathcal{P}_i^-} \exp\Big(\frac{\mathrm{sim}(\hat{p}_i, \hat{p}_k)}{\tau}\Big)},
\end{equation}

where $\tau$ is a temperature hyperparameter, $\mathcal{P}_i^+$ denotes same-class prototypes, and $\mathcal{P}_i^-$ denotes different-class prototypes. 

The overall loss is averaged over all prototypes:
\begin{equation}
\mathcal{L}_{\text{HyperPAS}} = \frac{1}{K} \sum_{i=1}^{K} \mathcal{L}_i.
\end{equation}

This formulation promotes inter-class separation while preserving class-specific directions in hyperbolic space.


The final training objective is a weighted combination of all terms:
\begin{equation}
\mathcal{L} = \mathcal{L}_{\text{CE}} 
+ \lambda_1 \mathcal{L}_{\ell_1} 
+ \lambda_2 \mathcal{L}_{\text{trans}} 
+ \lambda_3 \mathcal{L}_{\text{MAE}} 
+ \lambda_4 \mathcal{L}_{\text{cls/sepn}} 
+ \lambda_5 \mathcal{L}_{\text{HyperPAS}},
\end{equation}
where $\lambda_1, \dots, \lambda_5$ are hyperparameters controlling the weight of each term.

\section{Experiments and Results}
\subsection{Dataset and Implementation}
We evaluate our method on AP4 videos from a private echo dataset collected with institutional approval from a tertiary care center. 
In total, the dataset contains 141,086 unique cines from 32,818 studies. LVFP labels were derived from clinical text reports, yielding a class distribution of 85\% normal and 15\% elevated. Data were split at the study level into training, validation, and test sets using a 70:15:15 ratio to prevent patient-level leakage.

Experiments use fixed-length 32-frame video clips (~one cardiac cycle) with precomputed, frozen DINOv3 features of size $16 \times 14 \times 14 \times 512$~\cite{simeoni2025dinov3}. We employ 10 prototypes per class, with initial $E/e'$ values uniformly distributed between 1 and 25. Optimization runs for 31 epochs with batch size 32, using a learning rate of $1\times10^{-3}$ for the $E/e'$ head and $1\times10^{-4}$ for all other parameters. Training uses class-balanced sampling, and loss weights largely follow ProtoASNet~\cite{vaseli2023protoasnet}, with additional weights of 0.2 for $\mathcal{L}_{\text{MAE}}$ and 0.1 for $\mathcal{L}_{\text{HyperPAS}}$.

\subsection{Evaluation}


\textbf{Quantitative Results.} Tables~\ref{tab:cine_results} and~\ref{tab:study_results} summarize quantitative results at cine- and study-levels. We compare HypOProto against four baselines: EchoPrime + MLP~\cite{vukadinovic2024echoprime}, a linear classifier on frozen EchoPrime embeddings; DINO + MLP~\cite{simeoni2025dinov3}, a linear classifier on frozen DINO embeddings; Proto + DINO~\cite{simeoni2025dinov3,vaseli2023protoasnet}, a Euclidean prototypical network using the same DINO embeddings; and our implementation of the fully trained closed-source Akerman et al.~\cite{akerman2023automated} model.

At the cine level, HypOProto achieves the best performance across all baselines. The most notable improvement is in the elevated LVFP class (F1 0.63), indicating that the hyperbolic ordinal formulation better handles elevated cases, which tend to be closer to the diagnostic boundary. 

At the study level, computed as the mean of the softmax probabilities from cine-level logits, HypOProto attains the best mean and class-wise F1 for normal and elevated LVFP. While EchoPrime achieves the highest balanced accuracy, it lacks intrinsic interpretability. It also surpasses the Akerman et al.\cite{akerman2023automated} model, while requiring substantially less training effort due to frozen features.









\begin{table*}[t]
\caption{Cine-level (N = 21,120) balanced accuracy and F1 scores reported on the test set. Results are reported as mean (stdev) over three runs, with best results in bold.}
\label{tab:cine_results}
\centering
\begin{tabular}{l|cccc|c|c}
\multirow{2}{*}{Method} 
& \multicolumn{4}{c|}{Cine-level}
& \multirow{2}{*}{E2E} 
& \multirow{2}{*}{XAI}
\\ 
& bACC$\uparrow$ 
& Avg F1 $\uparrow$ 
& Normal F1$\uparrow$ 
& Elevated F1 $\uparrow$ 
& \\
\hline 
\hline

EchoPrime+MLP~\cite{vukadinovic2024echoprime}
& $80.02(0.19)$ 
& $0.70 (0.01)$ 
& $0.87 (0.01)$ 
& $0.54 (0.01)$ 
& \xmark 
& \xmark \\

DINO+MLP~\cite{simeoni2025dinov3} 
& $76.13 (0.20)$ 
& $0.67 (0.01)$ 
& $0.84 (0.01)$ 
& $0.50 (0.01)$ 
& \xmark 
& \xmark \\

Proto+DINO~\cite{simeoni2025dinov3,vaseli2023protoasnet}
& $79.45(0.46)$ 
& $0.72(0.00)$ 
& $0.88 (0.01)$ 
& $0.56 (0.01)$ 
& \xmark 
& \cmark \\

Akerman et al.~\cite{akerman2023automated}
& $79.65(0.53)$ 
& $0.71(0.01)$ 
& $0.87 (0.01)$ 
& $0.56 (0.01)$ 
& \cmark 
& \xmark \\

HypOProto (Ours)
& $\mathbf{82.57(0.43)}$ 
& $\mathbf{0.76(0.01)}$ 
& $\mathbf{0.91 (0.01)}$ 
& $\mathbf{0.63 (0.02)}$ 
& \xmark 
& \cmark \\

\end{tabular}
\end{table*}

\begin{table*}[t]
\caption{Study-level (N = 4,894) balanced accuracy and F1 scores reported on the test set. Results are reported as mean (stdev) over three runs, with best results in bold.}
\label{tab:study_results}
\centering
\begin{tabular}{l|cccc|c|c}
\multirow{2}{*}{Method} 
& \multicolumn{4}{c|}{Study-level}
& \multirow{2}{*}{E2E} 
& \multirow{2}{*}{XAI}
\\ 
& bACC$\uparrow$ 
& Avg F1 $\uparrow$ 
& Normal F1$\uparrow$ 
& Elevated F1 $\uparrow$ 
&
\\
\hline 
\hline

EchoPrime+MLP~\cite{vukadinovic2024echoprime} 
& $\mathbf{84.33 (0.40)}$ 
& $0.75 (0.01)$ 
& $0.88 (0.01)$ 
& $0.61 (0.01)$ 
& \xmark 
& \xmark \\

DINO+MLP~\cite{simeoni2025dinov3} 
& $80.70 (0.24)$ 
& $0.73 (0.02)$ 
& $0.88 (0.02)$ 
& $0.58 (0.02)$ 
& \xmark 
& \xmark \\

Proto+DINO~\cite{simeoni2025dinov3,vaseli2023protoasnet} 
& $81.49 (0.02)$ 
& $0.76 (0.00)$ 
& $0.90 (0.00)$ 
& $0.61 (0.00)$ 
& \xmark 
& \cmark \\

Akerman et al.~\cite{akerman2023automated}
& $82.60 (0.58)$ 
& $0.75 (0.01)$ 
& $0.89 (0.01)$ 
& $0.61 (0.01)$ 
& \cmark 
& \xmark \\

HypOProto (Ours)
& $83.97(0.15)$ 
& $\mathbf{0.79(0.01)}$ 
& $\mathbf{0.91 (0.01)}$ 
& $\mathbf{0.66 (0.01)}$ 
& \xmark 
& \cmark \\

\end{tabular}
\end{table*}

\begin{figure}[t]
\centering
\includegraphics[width=0.99\textwidth]{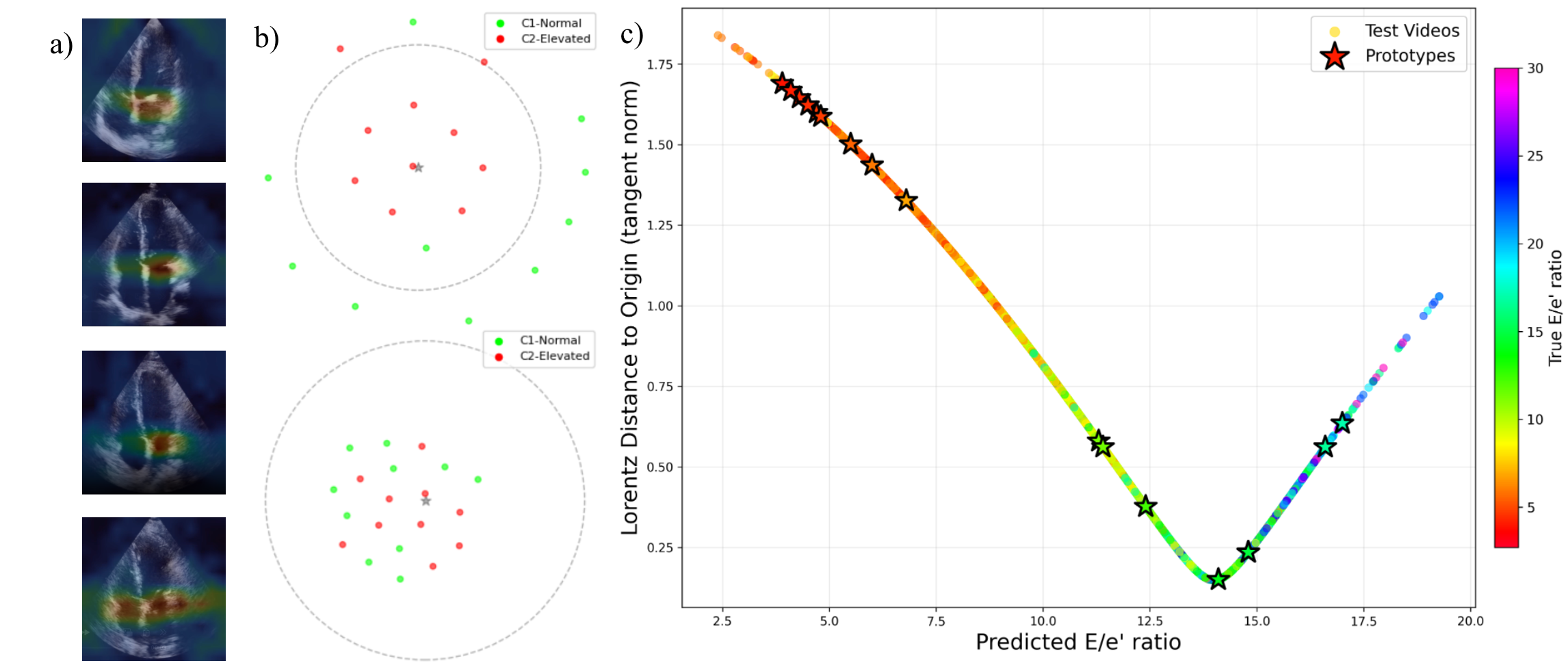}
\caption{a) Visualizations of learned prototypes. b) CoSNE plot of HypOProto (top) \emph{vs.} without radially enforced ordinality (bottom). c) Hyperboloid projections of videos and prototypes, colored by ground-truth $E/e'$ to illustrate preserved ordinal structure.}
\label{fig:qual}
\end{figure}
\subsubsection{Qualitative Results.}
Qualitative visualizations (Figure~\ref{fig:qual}) support that the model learns clinically meaningful and consistent representations. Prototype activation maps (a) show concentrated attention around the mitral valve inflow region, demonstrating anatomically aligned and physiologically grounded explanations. In the CoSNE projection (center-top)~\cite{guo2022co}, most elevated prototypes lie closer to the hyperbolic root while normal prototypes extend farther along the manifold. This reflects the dataset distribution where elevated studies cluster near the diagnostic threshold with a mean $E/e'$ of 17, while normal studies lie farther below it with a mean of 8, making the elevated group harder to separate. The normal prototypes closer to the root correspond to borderline normal cases, while a few elevated prototypes farther out represent more extreme elevated pathology, demonstrating that the geometry captures the full ordinal spectrum rather than enforcing a rigid binary split. In contrast, the CoSNE embedding without $E/e'$-based ordinality (center-bottom) shows substantial mixing between normal and elevated prototypes, resulting in a less interpretable and separable representation.  
The hyperboloid plot (right) shows that both videos and prototypes are positioned along the manifold according to the predicted $E/e'$ value that determines hyperbolic distance from the root. The color map reflects the ground-truth $E/e'$, and its smooth alignment with radial position indicates that the prediction head preserves the correct ordinal ordering. 
Notably, the dataset contains substantially more normal than elevated studies, with most normal cases far from the threshold and elevated cases clustered near it. The hyperbolic formulation accommodates this imbalance by allocating greater angular separation near the root, where borderline and mildly elevated cases concentrate.

\subsubsection{Ablation Study.}
\begin{table}[t]
\caption{Study-level ablation on the validation set (N = 4,922). Results are reported over three runs, with the best performance in \textbf{bold} and the second best \underline{underlined}. Each experiment removes key hyperbolic and loss components of the proposed model.}
\label{tab:ablation}
\centering
\begin{tabular}{lc|cc|cc}
\multicolumn{2}{c|}{\multirow{2}{*}{Method}} 
& \multicolumn{2}{c|}{Study-level}   & {\multirow{2}{*}{Sparsity $\downarrow$}}  & {\multirow{2}{*}{Diversity $\uparrow$}}     \\
\multicolumn{2}{l|}{} 
& bACC $\uparrow$ & Avg F1 $\uparrow$  & &  \\ \hline \hline
 & Proto+Dino          
& 82.05(0.77)       & 0.75(0.01)     & 0.68(0.02) & \textbf{0.97(0.03)}         \\
 & w/o Ord            
& 83.31(1.53)       & \textbf{0.78(0.02)}    & \textbf{0.53(0.09)} & 0.52(0.16)         \\
 & w/o $\mathcal{L}_\text{HyperPAS}$                
& \textbf{84.04(0.52)}       & \underline{0.77(0.01)}     & 0.63(0.01) & 0.63(0.03)         \\
 & All parts (ours)              
& \underline{83.89(0.19)}      & \textbf{0.78(0.01)}   & \underline{0.61(0.01)} & \underline{0.70(0.09)}        
\end{tabular}
\end{table}

The ablation results show that incorporating hyperbolic ordinality and the HyperPAS loss improves prototype quality while maintaining competitive performance. The full model achieves the best average F1 and second-best balanced accuracy at the study level, indicating a strong balance between sensitivity and overall classification performance. Removing the ordinality component eliminates the predicted $E/e'$ regression head; instead, features are projected to the hyperboloid using the norm of ${Z}$ to determine the radius, following the strategy used in~\cite{vaseli2025hiprotonet}. This slightly decreases balanced accuracy and prototype diversity, and removes the clinically grounded interpretation of the radial dimension, decoupling hyperbolic geometry from severity information. As a result, the embedding space becomes less structured, as shown in Fig.~\ref{fig:qual} (center). Omitting $\mathcal{L}_\text{HyperPAS}$ yields slightly increased balanced accuracy but produces less sparse and more redundant prototypes. This indicates that angular regularization promotes directional diversity in the tangent space. Compared with the Euclidean Proto+DINO baseline, all hyperbolic variants improve classification metrics. Overall, the proposed formulation attains top-tier predictive performance while consistently promoting sparser and more diverse prototypes.

\subsection{Discussion}

\subsubsection{Failure Analysis.}We stratified a subset of 18,677 test videos by whether the LVFP label agreed with the ground-truth $E/e'$ threshold of 14. Concordant cases, where both signals aligned, achieved high accuracy (87.9\% for normal with $E/e' < 14$, n=16,079; 83.3\% for elevated with $E/e' > 14$, n=2,062). Discordant cases performed substantially worse (59.7\% for elevated with $E/e' < 14$, n=362; 31.6\% for normal with $E/e' > 14$, n=174). These cases likely reflect label noise or patients whose physiology deviates from standard guidelines.

\subsubsection{Conclusion.}To address the challenges of interpretable LVFP classification from echo, we proposed HypOProto, a hyperbolic, ordinal prototype-based framework built on a frozen, explainable patch-based foundation model. By arranging prototypes along the physiological $E/e'$ scale, the model encodes clinically meaningful ordinal structure, placing borderline cases near the hyperboloid root and clearly normal or elevated cases outward to reflect diagnostic certainty. Additionally, visualizations highlight anatomically relevant regions.
Future work will focus on these cases and explore a third, indeterminate class—not present in this dataset—to capture patients who are neither normal nor elevated due to diverse physiological factors.
\bibliographystyle{splncs04}
\bibliography{ref}

@article{vukadinovic2024echoprime,
  title={Comprehensive echocardiogram evaluation with view primed vision language AI},
  author={Vukadinovic, Milos and Chiu, I-Min and Tang, Xiu and Yuan, Neal and Chen, Tien-Yu and Cheng, Paul and Li, Debiao and Cheng, Susan and He, Bryan and Ouyang, David},
  journal={Nature},
  pages={1--3},
  year={2025},
  publisher={Nature Publishing Group UK London}
}

@inproceedings{vaseli2023protoasnet,
  title={Protoasnet: Dynamic prototypes for inherently interpretable and uncertainty-aware aortic stenosis classification in echocardiography},
  author={Vaseli, Hooman and Gu, Ang Nan and Ahmadi Amiri, S Neda and Tsang, Michael Y and Fung, Andrea and Kondori, Nima and Saadat, Armin and Abolmaesumi, Purang and Tsang, Teresa SM},
  booktitle={International conference on medical image computing and computer-assisted intervention},
  pages={368--378},
  year={2023},
  organization={Springer}
}

@inproceedings{vaseli2025hiprotonet,
  title={HiProtoNet: Hyperbolic Hierarchy-Aware Part Prototypes for Aortic Stenosis Severity Classification},
  author={Vaseli, Hooman and Wu, Victoria and Kim, Diane and Tsang, Michael Y and Gu, Ang Nan and Luong, Christina and Abolmaesumi, Purang and Tsang, Teresa SM},
  booktitle={International Workshop on Advances in Simplifying Medical Ultrasound},
  pages={197--207},
  year={2025},
  organization={Springer}
}

@inproceedings{ghamary2025protoefnet,
  title={ProtoEFNet: Dynamic Prototype Learning for Inherently Interpretable Ejection Fraction Estimation in Echocardiography},
  author={Ghamary, Yeganeh and Wu, Victoria and Vaseli, Hooman and Luong, Christina and Tsang, Teresa and Bigdeli, Siavash A and Abolmaesumi, Purang},
  booktitle={International Workshop on Interpretability of Machine Intelligence in Medical Image Computing},
  pages={149--159},
  year={2025},
  organization={Springer}
}

@article{akerman2023automated,
  title={Automated echocardiographic detection of heart failure with preserved ejection fraction using artificial intelligence},
  author={Akerman, Ashley P and Porumb, Mihaela and Scott, Christopher G and Beqiri, Arian and Chartsias, Agisilaos and Ryu, Alexander J and Hawkes, William and Huntley, Geoffrey D and Arystan, Ayana Z and Kane, Garvan C and others},
  journal={JACC: Advances},
  volume={2},
  number={6},
  pages={100452},
  year={2023},
  publisher={American College of Cardiology Foundation Washington DC}
}

@article{hu2024hyperbolic,
  title={Hyperbolic geometry-driven robustness enhancement for rare skin disease diagnosis},
  author={Hu, Yang and Chen, Yuanyuan and Xing, Xiaohan and Zhang, Jingfeng and Yerzhanuly, Bolysbek Murat and Matkerim, Bazargul and Xia, Yong},
  journal={IEEE Journal of Biomedical and Health Informatics},
  volume={29},
  number={3},
  pages={2161--2171},
  year={2024},
  publisher={IEEE}
}

@inproceedings{gonzalez2025hyperbolic,
  title={Is Hyperbolic Space All You Need for Medical Anomaly Detection?},
  author={Gonzalez-Jimenez, Alvaro and Lionetti, Simone and Amruthalingam, Ludovic and Gottfrois, Philippe and Gr{\"o}ger, Fabian and Pouly, Marc and Navarini, Alexander A},
  booktitle={International Conference on Medical Image Computing and Computer-Assisted Intervention},
  pages={312--322},
  year={2025},
  organization={Springer}
}

@inproceedings{khrulkov2020hyperbolic,
  title={Hyperbolic image embeddings},
  author={Khrulkov, Valentin and Mirvakhabova, Leyla and Ustinova, Evgeniya and Oseledets, Ivan and Lempitsky, Victor},
  booktitle={Proceedings of the IEEE/CVF conference on computer vision and pattern recognition},
  pages={6418--6428},
  year={2020}
}

@inproceedings{vaseli2025happi,
  title={HAPPI: Hyperbolic Hierarchical Part Prototypes for Image Recognition},
  author={Vaseli, Hooman and Wu, Victoria and Kondori, Nima and To, Nguyen Nhat Minh and Fung, Andrea and Gu, Ang Nan and Abolmaesumi, Purang},
  booktitle={Proceedings of the IEEE/CVF International Conference on Computer Vision},
  pages={685--694},
  year={2025}
}

@article{simeoni2025dinov3,
  title={Dinov3},
  author={Sim{\'e}oni, Oriane and Vo, Huy V and Seitzer, Maximilian and Baldassarre, Federico and Oquab, Maxime and Jose, Cijo and Khalidov, Vasil and Szafraniec, Marc and Yi, Seungeun and Ramamonjisoa, Micha{\"e}l and others},
  journal={arXiv preprint arXiv:2508.10104},
  year={2025}
}

@inproceedings{law2019lorentzian,
  title={Lorentzian distance learning for hyperbolic representations},
  author={Law, Marc and Liao, Renjie and Snell, Jake and Zemel, Richard},
  booktitle={International conference on machine learning},
  pages={3672--3681},
  year={2019},
  organization={PMLR}
}

@inproceedings{huang2025hyperpath,
  title={HyperPath: Knowledge-Guided Hyperbolic Semantic Hierarchy Modeling for WSI Analysis},
  author={Huang, Peixiang and Huang, Yanyan and Zhao, Weiqin and He, Junjun and Yu, Lequan},
  booktitle={International Conference on Medical Image Computing and Computer-Assisted Intervention},
  pages={262--272},
  year={2025},
  organization={Springer}
}

@inproceedings{desai2023hyperbolic,
  title={Hyperbolic image-text representations},
  author={Desai, Karan and Nickel, Maximilian and Rajpurohit, Tanmay and Johnson, Justin and Vedantam, Shanmukha Ramakrishna},
  booktitle={International Conference on Machine Learning},
  pages={7694--7731},
  year={2023},
  organization={PMLR}
}

@misc{ase2025diastolic,
  title        = {Recommendations for the Evaluation of Left Ventricular Diastolic Function by Echocardiography},
  author       = {{American Society of Echocardiography}},
  year         = {2025},
  month        = {July},
  howpublished = {\url{https://www.asecho.org/wp-content/uploads/2025/07/Left-Ventricular-Diastolic-Function.pdf}},
  note         = {ASE guideline document}
}

@article{minners2010inconsistent,
  title={Inconsistent grading of aortic valve stenosis by current guidelines},
  author={Minners, Jan and Allgeier, Martin and Gohlke-Baerwolf, Christa and Kienzle, Rolf-Peter and Neumann, Franz-Josef and Jander, Nikolaus},
  journal={Heart},
  volume={96},
  number={18},
  pages={1463--1468},
  year={2010}
}

@inproceedings{guo2022co,
  title={Co-sne: Dimensionality reduction and visualization for hyperbolic data},
  author={Guo, Yunhui and Guo, Haoran and Yu, Stella X},
  booktitle={Proceedings of the IEEE/CVF Conference on Computer Vision and Pattern Recognition},
  pages={21--30},
  year={2022}
}
\end{document}